\documentclass[conference]{IEEEtran}
\IEEEoverridecommandlockouts
\usepackage{cite}
\usepackage{amsmath,amssymb,amsfonts}
\usepackage{algorithmic}
\usepackage{graphicx}
\graphicspath{{figures/}}
\usepackage{textcomp}
\usepackage{xcolor}
\usepackage{float}
\usepackage{tikz}
\usepackage{tabularx}
\usetikzlibrary{shapes, arrows, positioning}
\usepackage[T1]{fontenc}
\usepackage[hidelinks]{hyperref}

\def\BibTeX{{\rm B\kern-.05em{\sc i\kern-.025em b}\kern-.08em
    T\kern-.1667em\lower.7ex\hbox{E}\kern-.125emX}}
\begin{document}
\newcommand{\todo}[1]{{\color{red}#1}}

\title{Augmented Kinesthetic Teaching: Enhancing Task Execution Efficiency through Intuitive Human Instructions}

\author{Cheng Tang$^{1}$, Jiaming Zhong$^{2}$, Yue Hu$^{2}$
\thanks{$^{1}$ Department of Electrical and Computer Engineering, University of Waterloo, Canada {\tt\footnotesize cheng.tang@uwaterloo.ca}}
\thanks{$^{2}$ Department of Mechanical and Mechatronics Engineering, University of Waterloo, Canada}
\thanks{We acknowledge the support of the Natural Sciences and Engineering Research Council of Canada (NSERC), funding reference number RGPIN-2022-03857.}
}

\maketitle

\begin{abstract}

In this paper, we present a complete and efficient implementation of a knowledge-sharing augmented kinesthetic teaching approach for efficient task execution in robotics. Our augmented kinesthetic teaching method integrates intuitive human feedback, including verbal, gesture, gaze, and physical guidance, to facilitate the extraction of multiple layers of task information including control type, attention direction, input and output type, action state change trigger, etc., enhancing the adaptability and autonomy of robots during task execution. We propose an efficient Programming by Demonstration (PbD) framework for users with limited technical experience to teach the robot in an intuitive manner. The proposed framework provides an interface for such users to teach customized tasks using high-level commands, with the goal of achieving a smoother teaching experience and task execution. This is demonstrated with the sample task of pouring water.

\end{abstract}


\section{Introduction}

As industries continue to embrace automation and robotics to streamline processes and amplify productivity, the need for robots that can genuinely collaborate with humans becomes increasingly pronounced. Despite the considerable advancements made in robotics, the realm of human-robot interaction remains a significant challenge. 

Several studies have explored frameworks that enable robots to learn from human demonstrations through kinesthetic teaching or by segmenting observations.

Frameworks that optimize planning time by simplifying tasks into a rooted tree structure and identifying steps from human narration, utilizing hierarchical structures have been proposed by Saveriano et al. \cite{saveriano2019} and Moheseni-Kabir et al.\cite{mohseni-kabir2017}. Meanwhile, the focus of Willibald et al. \cite{willibald2020, willibald2022} has been on collaborative robot programming techniques, emphasizing anomaly detection for user adjustments and segmenting demonstrations into simpler skills in a task graph. Cruz et al. \cite{cruz2020} and Caccavale et al. \cite{caccavale2019}, \cite{caccavale2018} have ventured into the realm of adaptive robot systems, with Cruz exploring dynamic adjustments using Reconfigurable Behavior Trees (RBTs) and Caccavale emphasizing adaptive attention mechanisms combined with physical human-robot interaction. Agostini et al. \cite{agostini2020} tackled the challenges of geometric and symbolic reasoning in robot movements, emphasizing motion dependencies across actions. Finally, advanced robot learning skills have made significant strides in recent research, where Zanchettin et al. \cite{zanchettin2023} emphasized broader skill generalizations and Kormushev et al.\cite{kormushev2012} focused on tasks involving both position and force.

Gopika et al.'s \cite{Gopika2023} study has done preliminary exploration on the incorporation of human feedback for kinesthetic teaching. They highlighted the potential benefits of incorporating multimodal cues, such as gaze and verbal descriptions, into kinesthetic teaching. Their results suggest that multimodal cues can provide deeper insights into a programmer's intent, challenges faced during teaching, and a more holistic understanding of the task being demonstrated. Such an approach, which fuses multimodal cues like speech and gaze with traditional Program by Demonstration (PbD), promises a richer robot learning experience and can possibly lead to improved robot-human collaborative outcomes \cite{Gopika2023}. While the study provided valuable insights into human feedback and its impact on robot learning, there remains a need to explore the efficient means of implementation for augmented kinesthetic teaching to reach an integrated and intuitive solution.

The functional object-oriented network (FOON) developed by David Paulius et al. (2021)\cite{Paulius2021}, was a bipartite graph representation designed to represent a robot's understanding of its environment and tasks. This work proposed a roadmap for developing a generalizable structured task planning framework, enhancing effective robot knowledge acquisition from human demonstrations. 

In this paper, we propose the implementation of an augmented kinesthetic teaching pipeline that integrates multimodal intuitive human feedback during the task execution process which allows \textit{Learning by Demonstration} from very few examples by utilizing \textit{knowledge-sharing control} \footnote{Available at \href{https://github.com/hushrilab/augmented-kinesthetic-teaching}{https://github.com/hushrilab/augmented-kinesthetic-teaching}}. Where with intuitive we refer to an interaction pipeline that is similar to the ones that occur between humans. Through semantic-based subspace selection and regression reasoning, the collected prior knowledge can be efficiently transferred to new tasks to achieve adaptable and scalable interactions. 
By incorporating an object-oriented approach and leveraging automating the labeling process in machine learning from classical classification, regression tasks to deep learning applied in Computer Vision (CV), and Natural Language Processing (NLP) techniques, we developed a practical and efficient interface. This interface aims at interpreting high-level commands including verbal and visual cues, gestures, and physical guidance, without requiring users to learn specific robot-centric instructions, targeting technological innovation and usability, aiming at making human-robot interaction more approachable. We utilize pouring water as a detailed demonstration use case showcasing the workflow from learning aspects to the execution of a learned task. 

\section{System Structure}

The implementation of the framework detailed in this section is available for public access at \cite{augmentedkinesthetic}. 

\subsection{Multimodal task selection interface}

Our system integrates an object-oriented representation of the environment, where individual objects are symbolized as unique classes. Each class of objects is associated with a set of actions. Registration of objects and actions is done by the user through intuitive interactions using the interface introduced below. Interaction recordings that contain the processed environment state and user input information and selected object and action then form a dataset for the machine learning model that predicts the most probable target object and action to be performed, namely the target prediction model. Verbal communications, visual cues, gestures, and physical interactions are processed and interpreted using deep learning models to form a portion of the input space of the training data used in the target prediction model. Once trained, the model predicts outcomes, enabling the robot to adapt to evolving situations without the human teacher having to cover every possible situation explicitly. Upon registration of an object, the robot initiates a learning process that involves tracking and labeling new objects, combined with user feedback, to fine-tune detection models for future customized inferences.

In detail, every action a robot can perform is broken down into smaller sequences. These sequences can be represented as individual "steps" or "boxes" that guide the robot. Each box is called a state control box. It acts like a segment in a detailed instruction manual. Moving from one box to another requires a trigger - a specific condition or a set of conditions. These are 'state change triggers'. They signal the robot when to transition from one step of the action to the next. Each state control box is associated with two parameters: its input type and its output type. The input type (like visual data from a camera or pressure data from a sensor) informs the box about the current scenario. In contrast, the output type (such as torque applied to a joint) determines the physical response of the robot. These paired inputs and outputs dictate how the robot processes information and reacts in any given situation.

\begin{figure}[!htb]
\centering
\includegraphics[width=0.45 \textwidth]{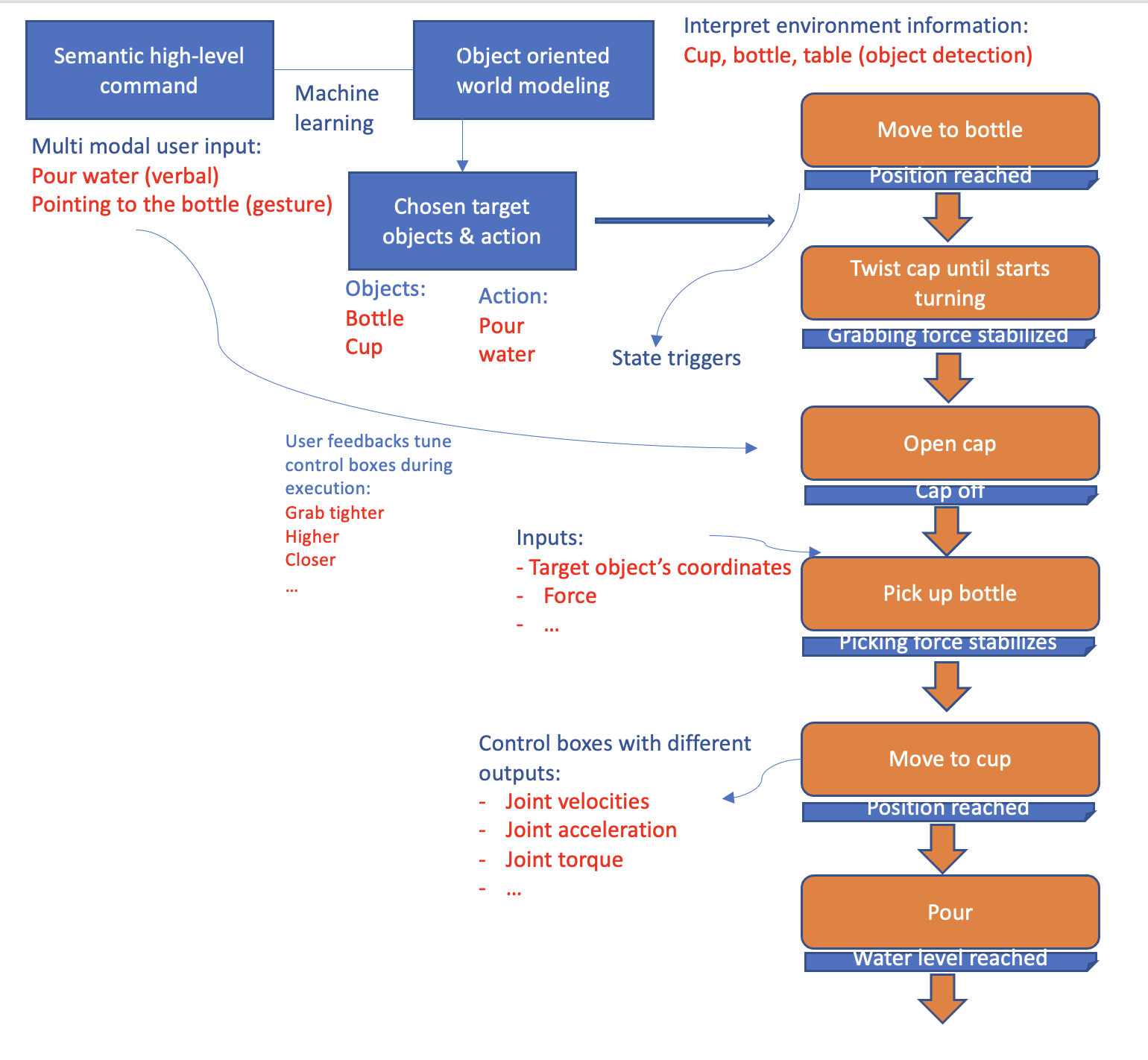}
\caption{\label{fig:control-architecture}Pouring water use case demonstration in multi-modal interface architecture. The orange boxes represent the control boxes. The text in red indicate elements specific to the pouring water action.}
\end{figure}

Figure \ref{fig:control-architecture} shows the interfacing pipeline for an example action pouring water, which will be discussed in detail below. 

\subsubsection{Object-Oriented World Representation}

Central to our approach is an object-oriented representation of the world. Objects within the environment are symbolized as individual classes. These classes encapsulate the associated methods (actions) that are applicable to the object. This encapsulation ensures that every interaction is inherently context-aware and that the robot's responses are specific to the object in question. This section presents a framework that integrates OOP principles with machine learning, enabling robots to learn and adapt from interactions within their environment using very few examples.

The following data structure describes the relationships in the context of object-oriented representation, the state control box will be discussed in section \ref{sec:knowledgesharing}: Knowledge sharing control.

\begin{small}
\begin{verbatim}
Class World:
    - objects: List[EnvironmentObject]
    + register_object(semantic_input)

Class EnvironmentObject:
    - ID
    - name
    - actions: List[Action]
    + register_action(semantic_input)

Class Action:
    - ID
    - name
    - procedure_flow: List[state_control_box]
    + register_state_control(semantic_input)

Class StateControlBox:
    - ID
    - name
    - input type
    - output type
    - state_change_trigger
    - input_output_mapper/state_controller
    
\end{verbatim}
\end{small}

\subsubsection{Recording Interactions}

During the teaching session, the robot archives each instruction from the human teacher. The pre-interaction world state including detected objects, and user commands such as any recognized verbal words are structured, as input vectors, encapsulating the relationship between environmental features and the user's semantic directive. Post-interaction, the consequential target object and action taken are then labeled as the target output. This systematic archiving of interactions facilitates a rich dataset, which facilitates the training of the target prediction model. The goal is for the robot to effectively discern and predict the impact of user commands on its environment. Table one \ref{tab:interactionDataset} shows a few simplified training data examples.

\begin{table*}[h]
\centering
\caption{Sample Interaction Dataset}\label{tab:interactionDataset}
\begin{tabularx}{\textwidth}{|X|X|X|X|X|}
\hline
\textbf{Detected Objects} & \textbf{Verbal Words} & \textbf{Gesture} & \textbf{Object} & \textbf{Action} \\
\hline
\{Cup, Spoon\} & "Pick" & Pointing to cup & Cup & Pick up cup \\
\hline
\{Bottle,Cup\} & "Pour" & N/A & Bottle & Pour water \\
\hline
\{Knife, Bread, Butter\} & "Spread the butter" & N/A & Bread & Spread Butter \\
\hline
\{Laptop, Mouse\} & N/A & Tapping motion & Laptop & Turn on \\
\hline
\end{tabularx}
\end{table*}

\subsubsection{Verbal Communication}

Common keywords are predefined including "closer", "tighter", "up", "down", etc. These labeled words are associated with control commands that can be used during real-time human feedback during teaching processes. 

In addition, during learning, any new keywords captured from the user are registered with corresponding word embedding used in NLP models. Which is then used as input for the target prediction model. For instance, upon learning to pour a drink, the human teacher mentions words such as "pour", "water", or "drink" multiple times. The robot registers these words and looks up the word embedding tables, forming partially the input of the training data. After a few complete teaching iterations, the target prediction model is updated. Post training, when a verbal command like "pour a drink" is issued again, the target prediction model, with the output from voice recognition, predicts the registered action "pouring water" to be a probable output, reinforcing context-aware action generation.

\subsubsection{Gesture Recognition and Interpretation}

The system implements predefined gestures classification for common gestures related to the robot's operation and manipulation including opening/closing the gripper, twisting and turning, pointing to an object, lifting, etc. These labeled outputs are then used as partial input for the target selection model as well as interpreted as real-time human feedback during teaching processes. For instance, a pointing gesture towards a shelf is interpreted as an affirmative cue to fetch an item from that location; during the teaching process of twisting a bottle cap a gesture of pinching is interpreted as closing the gripper even more to ensure grabbing the cap firmly.

\subsubsection{Visual Cues}

World information is primarily obtained through vision. Specifically, environmental objects and their relationships are interpreted visually using detection models. Output from the detection model is then used partially as input for the target prediction model.

\subsubsection{Physical Interaction}

Similar to gestures, common physical guidance operations are also classified using a pre-trained model. Interpreting environment and human input ranges from guiding the robot arm to a certain position, adjusting its grip strength, to redirecting its movement path, and nudging, etc. Each of these physical interactions is mapped to a corresponding directive in the robot's operation protocol. To be specific, if a user nudges the robot towards a door, it discerns this physical cue as a directive to move in that direction. This ensures real-time adaptability and fosters dynamic, multi-modal user-robot interactions.

\subsubsection{Initiating the Learning Process}

To adapt to the user, the robot initiates the continuous learning process. For example, by triggering the visual object learning mode and providing a reference to the object in the camera or real world, the robot autonomously captures images of unknown objects, segments likely areas of interest, and then seeks human confirmation. Simultaneously, the robot logs its camera's position and orientation, ensuring that every viewpoint is cataloged, allowing the robot to independently analyze, categorize, and label the object within its internal database, thus creating a visual dataset using an object tracking model. If discrepancies or mis-labelings are identified by the user, corrective feedback is incorporated immediately. This 'co-learning' approach serves two purposes: it generates a dataset on-the-fly, and it allows users to customize the robot's knowledge. Over time, as the robot gathers more data and gets more feedback, it fine-tunes its detection model. This ensures that each robot's understanding of the world is tailored to the specific needs and preferences of its user, adding a layer of personalization to the system.

\begin{figure}[!htb]
\centering
\includegraphics[width=0.4\textwidth]{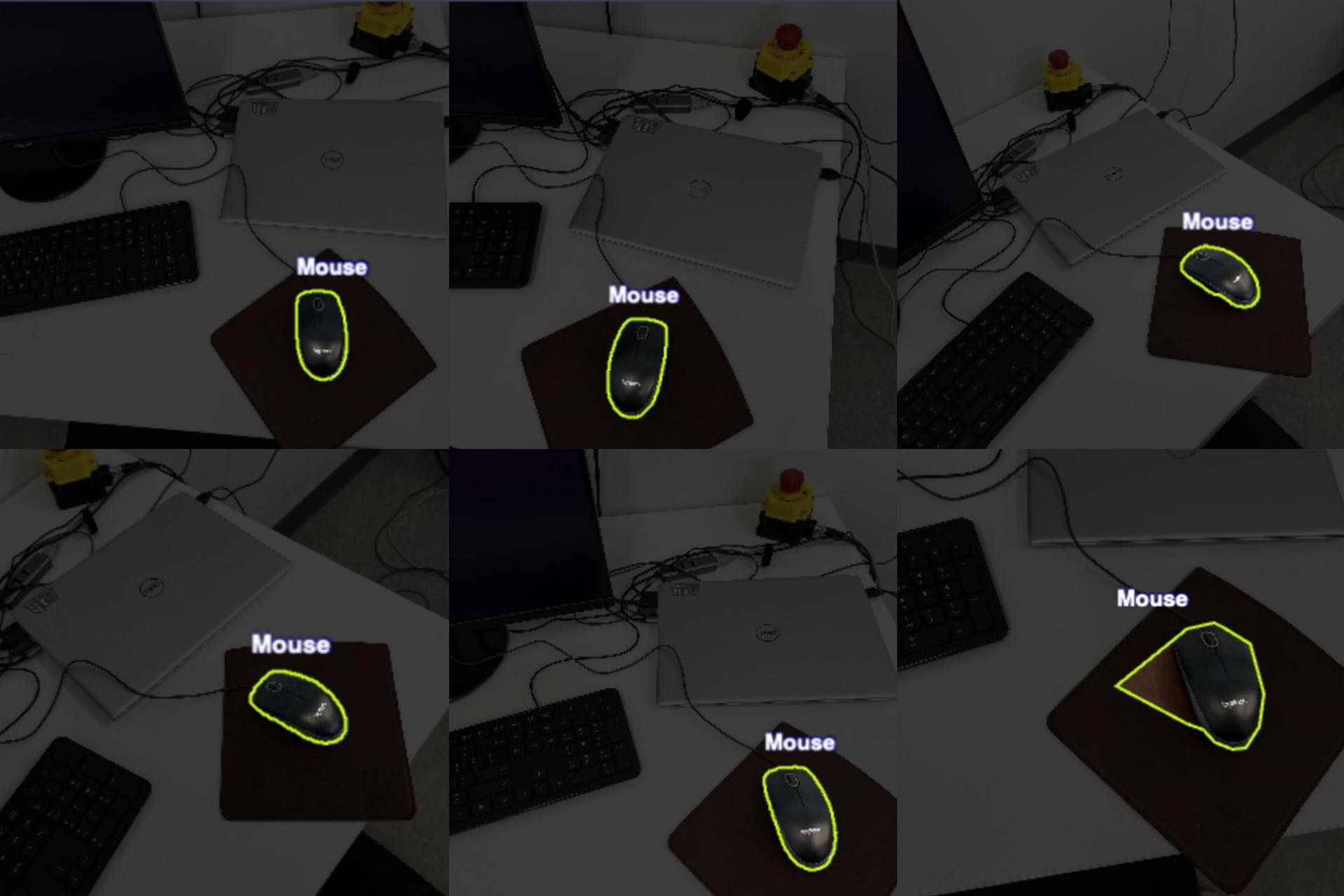}
\caption{\label{fig:registration}Object registration interface: prompting the user to select incorrect auto-labeled images and remove them from training data.}
\end{figure}

Figure \ref{fig:registration} shows the interfacing prompting the user to remove the incorrect labelled images.

\subsection{State control box definition}

A state control box is defined with an input, output, controller, and state change trigger, outlining the state control box's input-output type and their relationship as well as the criteria for exiting the state. During the teaching process, the robot interprets the human teacher's high-level command and decides on the input/output types on its own. 

\subsubsection{Input/Output Definition}
Upon hearing keywords or phrases such as "pay attention", "check", or "control", etc., the robot knows that the control input type is indicated. Keywords such as "grabbing force", "position", "location", "angle", or "speed" are classified in the pre-trained classification model, which predicts the state control box's input types. These keywords govern the information the robot must be aware of during an action execution, namely, what information or feedback should it care about. For instance, if the target object is the bottle, the verbal command "check your position" is received, the robot will change the control input to the relationship between its end effector pose and the bottle's pose, and control output as position control.

\subsubsection{Trajectory Definition}
Knowing the input and output type, the robot records and labels the trajectory autonomously. Trajectory information includes sensor recording from the corresponding input type, such as using a force-torque sensor or distance recorded using a camera and detection model, and control command with human feedback instruction. In the scenario of moving to the bottle, pose transformation between the end effector and bottle and position trajectory command are recorded during the teaching process. Input and output relationships are not limited in dimensionality. After a few iterations, the trajectories recorded have their input and output mapped and compared with the existing trajectories. A regressed version of a matched trajectory is then assigned to the corresponding state control box, outlining the input-output relationship.

\subsubsection{State Change Trigger Definition}
To mark a trigger as a transition signal between states, a similar user command interpretation approach is used. The robot determines the critical sensory input to be monitored upon hearing certain keywords. By triggering the state change, the robot records the threshold of the corresponding reading and marks it as a cue to exit that state control box.

\subsection{Knowledge sharing control}
\label{sec:knowledgesharing}

With the real world's repetitive nature, many tasks exhibit intrinsic similarities that can be leveraged for effective action execution. For example, opening a door or lifting an object, are tasks that follow a repetitive structure. Since their basic properties remain consistent across different scenarios, the core control strategy governing these actions can be reused. With shared knowledge, robots do not need to "learn from scratch" for every new yet similar task. Given the vast array of actions and corresponding strategies, the robot maintains a reservoir or 'pool' of control knowledge. 

Our system identifies tasks with similar intrinsic properties, which allows the sharing of basic underlying control strategies. This shared pool of control knowledge is systematically stored in various output spaces. While it might be labor-intensive to discern relationships between every pair of actions manually, our system turns to mathematical modeling. This ensures new actions are matched to existing ones efficiently. Multiple abstraction layers of models will be incorporated. The first layer reduces the dimensionality using the given new action input-output mapped trajectory. Following this is a regressor layer, which stores action knowledge connecting the input and output space. When a new action emerges, our system consults the classifier, followed by the regressor, to map inputs and trace the most analogous existing action trajectory.

\textbf{Subspace selection} The knowledge pool is composed of a dataset $D_n=\{ (\mathbf{X}_n, \mathbf{Y}_n, \mathbf{s}_n) \}$ that consists of all the data that can be collected, where $n$ is the number of data in the pool. $\mathbf{X}\in \mathbf{R}^p$ denotes the inputs, which usually consist of measured sensor signals or estimations, such as joint torque, pose, pressure, etc. $\mathbf{Y}\in \mathbf{R}^q$ denotes the output motion sequence needed for a task, which could be represented as high-level decisions or low-level trajectories and control commands. $\mathbf{s} \in \{ s_1, \ldots, s_k \}$ denotes the abstracted semantic labels, such as picking up, grabbing, lifting, etc. Ideally, the prior knowledge pool $D_n$ should include as much task data as possible that satisfies expectations.
However, not all dimensions are useful for a given type of semantic task $K \in \{1, \ldots, k\}$. On the contrary, too many irrelevant input dimensions will cause data sparseness and reduce learning efficiency. Therefore, finding a subspace $D_m=\{ (\mathbf{X}_m^r, \mathbf{Y}_m^q) | s = s_K \}$ that only containing valid input dimensions $r \leq p$ is necessary for the given semantic task $K$ before the regression inference.

Two types of common subspace selection methods are used. The first type is correlation analysis tools, such as the Pearson Correlation Coefficient (PCC) \cite{4389268}, which can easily find the most relevant input variables between the input $X$ and corresponding output $Y$. This method is simple and efficient and even can be processed online in real-time. However, the drawback is that multiple colinear dimensions could still exist in the subspace, which will suppress the accuracy of regression inference. The other type is the "model selection" criteria in statistical learning. The commonly used criteria can be roughly divided into three categories \cite{james2013introduction}: a) predictive risk function, e.g., Prediction Mean Squared Error (PMSE); b) coefficient of determination $R_{adj}^2$; c) likelihood-based information criteria: e.g., Akaike Information Criterion (AIC). The "forward-selection" method could find the smallest subset with the best criteria. 
It will start from the minimum subset and then add variables while comparing the above criteria for each selection to find the best one. Those criteria consider both data fitting and model complexity so over-fitting will be punished. 
One can choose any type of method to select the subspace $D_m$ according to the requirement of the task.

\textbf{Regression inference} Based on the selected subspace knowledge pool $D_m$, the efficiency and accuracy of the following regression inference process will be improved since the input space has been reduced from $n\times p$ to $m\times r$.

\begin{figure}[!htb]
\centering
\includegraphics[width=0.4\textwidth]{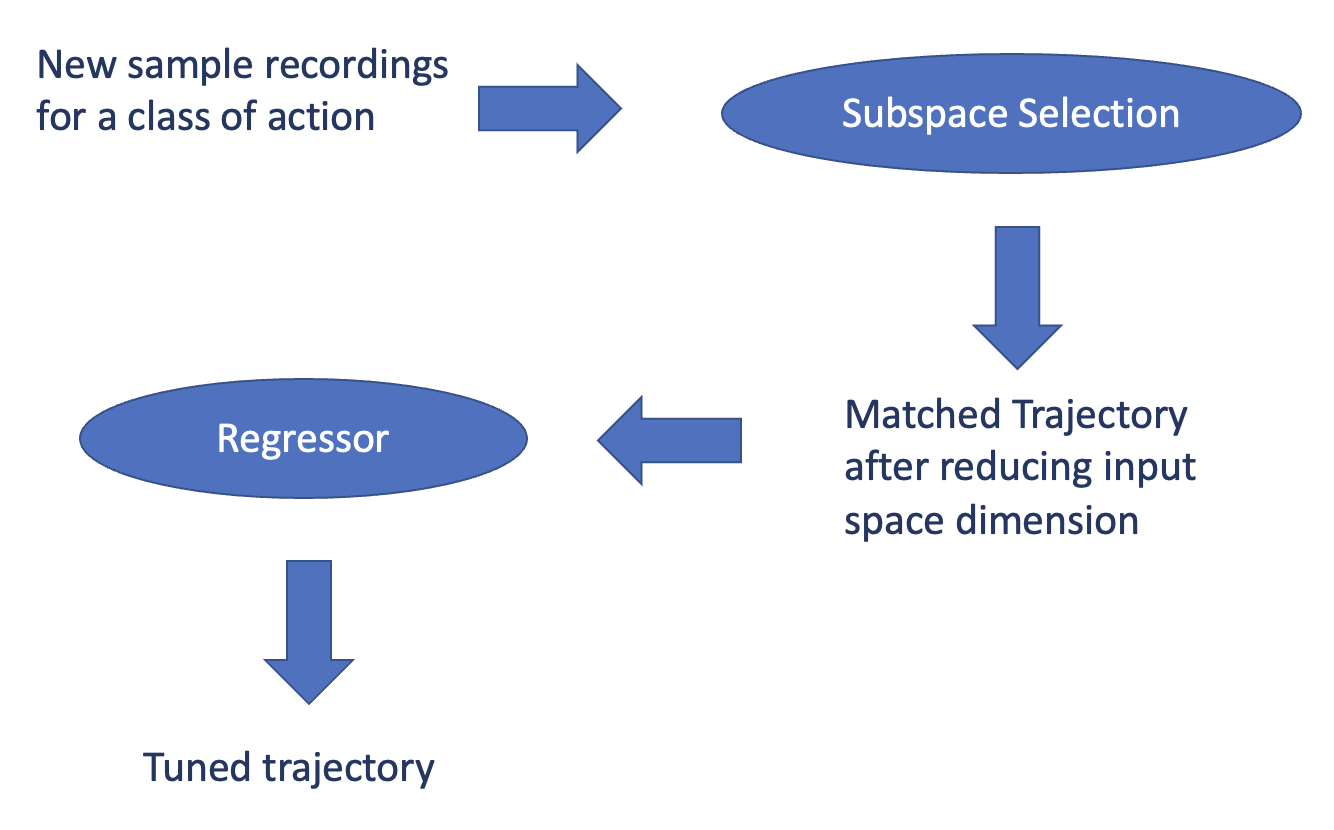}
\caption{\label{fig:trajectory}Trajectory matching and tuning flow chart.}
\end{figure}

Figure \ref{fig:trajectory} shows the workflow of trajectory matching.

A small number of real-world demonstrations help fine-tune the chosen strategy. These demonstrations allow the robot to adjust its existing trajectory, making it a closer fit to the new controlled trajectory. Post-refinement, the robot updates the particular state control box with the new, adjusted trajectory.

\section{Use Cases Demonstrations and Evaluation}
To demonstrate how our system integrates multi-modal inputs and object-oriented world modeling to learn and execute the task efficiently, we use the detailed example of the task of "pouring water" from a bottle. The following sections illustrate the detailed steps involved. The attached video illustrates the task in action.

\subsection{Object and Action Registration}

To initiate the task of "pouring water", the object "bottle" first is labeled and registered in the system by a human teacher. The bottle in the robot's camera frame is segmented, enabling the robot to track it. The object tracking model auto-labels the visual input as "bottle". Following this, "pouring water" is registered as an action affiliated with the bottle class. The sequence of actions required to achieve the pouring task is initiated based on the user’s semantic inputs and subsequently defined through a combination of visual cues, gestures, and other multi-modal interactions.

\subsection{Learning Registered Object by Auto-labelling}

The robot autonomously captures multiple images and angles of the bottle using its object tracking model where each frame is auto-labelled. To ensure accuracy and avoid incorrect inferences, the user is periodically prompted through the registration interface, to verify the labels. Incorrectly labeled images are then removed, ensuring that the robot does not incorporate them into its learning dataset. 

\subsection{Control Box Definition}

\subsubsection{Moving to the Bottle}
\textbf{Input/Output Definition}
Upon receiving verbal commands such as "move to it" and observing a gesture of pointing to the bottle, the robot perceives the bottle as its target. It sets its control input to be the relative position between the end effector and the bottle. The output is set as the trajectory the robot arm should follow to approach the bottle without any obstruction.

\textbf{State Change Trigger Definition}
The robot waits for cues such as "close enough" or a gesture that signals stop. Upon recognizing these cues, it records the current position as the ideal stopping point to pick up the bottle.

\textbf{User Feedback During Execution}
If the user notices the robot moving too fast or heading in a slightly wrong direction, feedback like "slowly", "to the left", or "higher" is provided. The robot would then adjust its output control command in real-time.

\subsubsection{Opening the Cap}
\textbf{Input/Output Definition}
Upon hearing cues such as "open it" and "be aware of force applied and orientation of the cap". The control input is then set by the pre-trained machine learning model to the gripping strength and the orientation of the cap relative to the end effector. The output is the twisting motion and gripper position required to open the cap.

\textbf{State Change Trigger Definition}
Once the cap starts to loosen, the robot feels a reduction in resistance. At this point, if the user says "That's enough" or provides a sign, the robot identifies this as the trigger to stop twisting and records the gripping force at the moment.

\textbf{User Feedback During Execution}
If the robot is exerting too much or too little force, the user might comment, "gentler" or "tighter". The robot modifies its grip or twisting strength accordingly.

\subsubsection{Picking Up the Bottle}
\textbf{Input/Output Definition}
Commands such as the verbal cue "lift it" and a gesture of lifting indicate that the robot should focus on position and gripping force. The control input focuses on the gripping strength and position, ensuring the bottle is securely held. The output defines the position trajectory the robot arm should follow as well as the gripping force.

\textbf{State Change Trigger Definition}
A phrase like "enough" or a gesture to stop would be the cue for the robot to cease the lifting action and maintain the bottle at its current height.

\textbf{User Feedback During Execution}
Feedback such as "firmer grip" or "more gently" helps the robot adjust its grip on the bottle. It remains adaptive to ensure the bottle is neither dropped nor crushed. 

\subsubsection{Pouring}
\textbf{Input/Output Definition}
With phrases like "rotate with an angle", the robot recognizes the need to tilt the end effector. The control input is the current tilt angle of the bottle, and the output is the end effector pose at which the robot should tilt to pour the liquid efficiently.

\textbf{State Change Trigger Definition}
Commands such as "stop pouring" or a hand gesture indicating stop will be the cues for the robot to stop further tilting movement. 

\textbf{User Feedback During Execution}
In case the liquid is pouring too quickly or too slowly, feedback like "faster" or "slower" would be used by the robot to adjust the pouring speed.

\subsubsection{Tightening the Cap}
\textbf{Input/Output Definition}
On hearing "seal it" or "close the cap", the robot prepares to reseal the bottle. The control input is the alignment of the cap with the bottle, and the output is the twisting motion to secure the cap in place.

\textbf{State Change Trigger Definition}
A decrease in twisting resistance or the user saying "That's tight enough" would be the cue for the robot to stop the action, ensuring the bottle is sealed without over-tightening.

\textbf{User Feedback During Execution}
If the robot is not aligning the cap properly or is twisting too hard, feedback like "align properly" or "ease up" can be provided. The robot adjusts its motion to correctly and efficiently seal the bottle. 

\subsection{Trajectory Recording}

As the robot performs each of the aforementioned tasks, it captures the trajectory. This data encompasses sensory information, such as grip strength when picking up the bottle or the exact angle and speed of tilt while pouring. User feedback directly modifies the control output using high-level instruction such as verbal cues, making the trajectory more accurate.

\subsection{Executing a Learned Task}

Once the robot has registered the object, defined its control boxes, recorded various trajectories, and identified the best ones, it is ready to execute the task autonomously. When prompted by the user, the robot, using its target prediction model, identifies the bottle, approaches it, and performs the pouring action based on the learned trajectory. Throughout this process, the robot remains receptive to user feedback, making real-time adjustments if necessary, ensuring the task's successful completion.

\subsection{Evaluation}

\subsubsection{Learning Time}

\paragraph{Time to Adoption}
The duration it takes for the robot to grasp and proficiently execute the pouring task is demonstrated in the video. This metric offers an understanding of how quickly the robot can adapt to new tasks, highlighting its ability to learn efficiently with our framework.

\paragraph{Trials to Achieve Success Rate}
The video also shows the number of trials the robot underwent before reaching a satisfactory success rate, Demonstrating the framework's consistency and persistence in refining the robot's skills through repetitive training.

\subsubsection{Success Rate in terms of Achieved Rate}
The final success rate, captured after numerous trials, is showcased in the video. 

\subsubsection{Adaptability in terms of Scenario-Based Performance}

The robot's adaptability is tested across diverse scenarios, each presenting its own set of challenges. The video illustrates how the robot manages these varying situations, with the success rates for each scenario being distinctly recorded. This provides a comprehensive view of the robot's flexibility and versatility in handling different contexts.

\subsubsection{Pouring Precision in terms of Accuracy and Neatness}
The precision of the robot's pouring action is assessed based on two primary factors: accuracy and neatness. While accuracy pertains to the robot's ability to pour the exact amount of liquid as intended, neatness evaluates how cleanly and orderly the action is performed. Any imprecision, such as spillages or erratic movements, is taken into account to offer a holistic view of the robot's performance in this specific task.

\section{Conclusion}

In this research, we introduced an approach to enhancing human-robot interaction using an augmented kinesthetic teaching method. We leveraged state-of-the-art NLP and CV to craft a multimodal interface that can interpret high-level human commands, including gestures and visual prompts, to direct robotic action without necessitating user familiarity with intricate robot-specific instructions.

Our system hinges on an object-oriented representation of the world. Each environmental object is encapsulated as a class, with associated actions or methods, ensuring the robot's interactions are contextually apt. Interactions are systematically archived, and this dataset informs the training of our machine-learning model, which aids the robot in understanding the implications of human directives on its environment.

Additionally, we've incorporated robust mechanisms for interpreting various communication modalities like verbal commands, gestures, visual cues, and even physical nudges. Each command, whether it's a word like "pour" or a gesture like pointing, is decoded and mapped to a corresponding action. Robots continuously refine their models by automatically collecting and labeling data with user feedback to enhance object recognition accuracy over time, thus adapting to user-specific needs and preferences through iterative interactions.

A novelty introduced in our system is knowledge sharing. Many tasks share inherent similarities, and our system recognizes and leverages these shared foundational control strategies, utilizing mathematical modeling. Furthermore, by segmenting actions into smaller sequences or state control boxes, we've crafted a modular approach to robotic instruction, where triggers inform transitions between sequential steps or actions.

\section{Future Work}

The system prioritizes ease and intuitiveness in its command structure, ensuring recognition and immediate response to user instructions, making it accessible to a broad user base irrespective of their technical background. To validate its user-centric design, future studies are planned to gather qualitative feedback and measure the system's practicality and user-friendliness. 

While our current system presents promising results, there are several aspects that can be tackled for future improvements, such as: considering richer semantic interpretations with further integration of NLP techniques; exploring methods for robots to share learnings to expedite the learning process; extending the Object Oriented classes with an adaptive mechanism to autonomously categorize new objects; integrate advanced safety mechanisms; incorporate user-specific customizations for robot behavior adjustments based on individual user preferences; and verify the intuitiveness with user experiments. Our next steps will aim at refining the system by addressing these challenges towards real-world applications.

\nocite{*}
\bibliographystyle{IEEEtran}
\bibliography{bib/IEEEabrv,bib/references}

\end{document}